\documentclass{article}

\PassOptionsToPackage{numbers}{natbib}
\usepackage[preprint]{nips_2018}

\usepackage[utf8]{inputenc} % allow utf-8 input
\usepackage[T1]{fontenc}    % use 8-bit T1 fonts
\usepackage{hyperref}       % hyperlinks
\usepackage{url}            % simple URL typesetting
\usepackage{booktabs}       % professional-quality tables
\usepackage{floatrow}
\usepackage{amsfonts}       % blackboard math symbols
\usepackage{nicefrac}       % compact symbols for 1/2, etc.
\usepackage{microtype}      % microtypography
\usepackage{float}
\usepackage{graphicx}
\usepackage[dvipsnames]{xcolor}
\usepackage{paralist,color}
\usepackage{multirow}
\usepackage{amsmath,bbm,amssymb}
\usepackage{algorithm}
\usepackage[noend]{algorithmic}
\usepackage{caption}
\usepackage{subcaption}
\usepackage{amsthm}

\newcommand{\bx}{\mathbf{x}}
\newcommand{\by}{\mathbf{y}}

\newcommand{\confounding}{confounding }

\newcommand{\confounded}{confounded }
\newcommand{\Confounding}{Confounding }

\newcommand{\xgems}{\textbf{xGEMs}}
\newcommand{\counterfactual}{manifold guided example}
\newcommand{\counterfactuals}{manifold guided examples}
\newcommand{\Counterfactual}{\textbf{xGEM}}
\newcommand{\Counterfactuals}{\xgems}
\mathchardef\mhyphen="2D

\newcommand{\vertiii}[1]{{\left\vert\kern-0.25ex\left\vert\kern-0.25ex\left\vert #1
    \right\vert\kern-0.25ex\right\vert\kern-0.25ex\right\vert}}

\def\bx{{\mathbf{x}}}
\def\by{{\mathbf{y}}}
\def\bz{{\mathbf{z}}}
\DeclareMathOperator*{\argmin}{arg\,min}
\DeclareMathOperator*{\argmax}{arg\,max}
\hypersetup{
    colorlinks,
    linkcolor={red!60!black},
    citecolor={blue!75!black},
    urlcolor={blue!80!red}
}

\theoremstyle{definition}
\newtheorem{definition}{Definition}[]

\title{xGEMs: Generating Examplars to Explain Black-Box Models}

\author{
Shalmali Joshi \\
UT Austin \\
 \texttt{shalmali@utexas.edu} %%\\
\AND
Oluwasanmi Koyejo \\
UIUC \\
 \texttt{sanmi@illinois.edu} \\
 \And
 Been Kim \\
 Google Brain \\
\texttt{beenkim@google.com} \\
\AND
Joydeep Ghosh \\
UT Austin \\
\texttt{jghosh@utexas.edu} %%\\
}

\begin{document}

\maketitle

\begin{abstract}
This work proposes \xgems \,: or \emph{manifold guided exemplars}, a framework to understand black-box classifier behavior by exploring the landscape of the underlying data manifold as data points cross decision boundaries.
To do so, we train an unsupervised implicit generative model -- treated as a proxy to the data manifold. We summarize black-box model behavior quantitatively by  perturbing  data samples  along the manifold. We demonstrate \xgems' ability to detect and quantify bias in model learning and also for understanding the changes in model behavior as training progresses.
\end{abstract}

\section{Introduction}

Machine learning algorithms have become widely deployed in domains beyond web based recommendation systems, like the criminal justice system~\citep{angwin2016machine}, clinical healthcare ~\citep{Callahan2017279} etc. For instance, risk assessment tools like COMPAS~\citep{angwin2016machine} produce learned recidivism scores to consequently determine the amount of pre-trial bail and detention. Similarly, medical interventions can impact health outcomes for patients, making institutions liable to provide explanations for their decisions. This has motivated regulatory agencies like the EU Parliament\footnote{in collaboration with the European Commission and the Council of the Eurpean Union} to codify a  right to data protection and ``obtain an explanation of the decision reached using such automated systems\footnote{https://www.privacy-regulation.eu/en/r71.htm}".

Systems that provide satisfactory explanations for the decisions of such learning algorithms have until recently been few and far between. It is challenging to characterize the specific nature of explanability mechanisms given their complexity and lack of consensus on the nature and sufficiency of such explanations~\citep{DoshiKim2017Interpretability, lipton2016mythos}. The problem is often compounded due to multiple levels of abstraction required to provide such explanations~\citep{miller2017explanation}. For instance, system level explanations as required by regulatory bodies are different from an abstraction that would assist practitioners of machine learning. This work focuses on providing explanations for low level understanding of model behavior, albeit at an abstraction beyond performance metrics. Such a suite of explanations not only help improve understanding of opaque models\footnote{\url{https://distill.pub/2018/building-blocks/}}~\citep{higgins2016early,karpathy2015visualizing} but can also uncover biases (inherent in the data) that models pick up on e.g. learned gender and racial biases~\citep{bolukbasi2016man}.

In this work, we posit that there need not be an inherent tradeoff between model performance and explanability, as is generally assumed (and found in \cite{kim2016examples, gupta2016monotonic,hughes2016supervised}). We propose an explanability tool that probes a supervised black-box model along the data manifold for explanations via examples and/or summaries. Demonstrating model behavior via examples is known to be beneficial for improving and understanding the decision making process~\citep{aamodt1994case}. Navigating the data manifold allows us to explore black-box behavior in different regions of the manifold. %Thus, we learn an approximate manifold of the data distribution using recent advances in implicit generative models~\citep{kingma2013auto,goodfellow2014generative}. We consider implicit generators as the class of stochastic models used to generate data points without prescribing a parametrization of the underlying data distribution. Thus we can explore model behavior in the range of the generator function. We note that this generative model is learned in an unsupervised way assuming access to the training samples (without label information) used to train the target black-box classifier. 
The proposed method can be utilized as a diagnostic tool to analyze training progression, compare classifier performance, and/or uncover inherent biases the classifier may have learned.
\section{Related Work}
Most closely related works to our approach are those that provide explanations by sub-selecting meaningful samples and/or semantically relevant features (like super-pixels) that highlight undesirable model behavior~\citep{NIPS2017_6993,kim2016examples}. Most of these methods require the selected samples to be part of training/test dataset. This means that if the training/test set did not include the instance that best explains a specific decision, we would have to settle for a suboptimal choice. Our method aims to relax this constraint by generating new examples that are better suited for this purpose. In terms of generating examples, adversarial criticisms~\citep{stock2017convnets} and the class of generative networks like GANs are relevant approaches. Specifically, ~\citep{stock2017convnets} use the adversarial attack paradigm as a means to select examples from existing training data to explain model behavior, similar to~\cite{kim2016examples}. 
%In other words, prior work on adversarial examples tend to focus on the worst case of the decision function. 
However note that the goal of generating adversarial examples and our explanations are fundamentally different. The primary goal of adversarial examples is to focus on exploiting the worst case confounding scenario given a decision boundary, while our work focuses on generating an example that lies on the data manifold as it crosses a decision boundary. See Figure~\ref{fig:simulated_data} for a more intuitive explanation.
%The goal of these examples are typically to change the prediction, rather than 
%On the other hand, it is well known that natural image data is more likely to lie in a low-dimensional data manifold within the ambient space. Adversarial attacks may therefore generate samples that may be less likely to fall on the manifold. 
We posit that it is important to uncover classifier behavior when data points are constrained to the data manifold. Such data instances are more `realistic' and likely to be created by the underlying phenomenon that led to the training data. They provide an alternative method to probe a black-box, specially in non-adversarial settings. They also characterize the residual vulnerabilities of a model that defends itself against adversarial attacks by detecting directed "noise" that is orthogonal to the manifold of the data or of an associated latent space. 

We position our work as a diagnostic framework for understanding model behavior at an abstraction that may be most useful to a data science practitioner and/or a machine learning expert. However, as suggested before, explainable models focus on different notions of explanability. For example,~\citet{koh2017understanding} use influence functions, motivated by robust statistics~\citet{cook1980characterizations} to determine importance of each training sample for model predictions.~\citet{li2015visualizing,selvaraju2016grad} focus on understanding the workings of different layers of a deep network and studying saliency maps for feature attribution~\citep{simonyan2013deep,smilkov2017smoothgrad,sundararajan2017axiomatic}. Saliency methods, while powerful, can be demonstrated to be unreliable without stronger conditions over the saliency model~\citep{2017arXiv171100867K,adebayo2018interpretation}. Other paradigms of explainable models focus on locally approximating complex models using a simpler functional form to approximate  the (local) decision boundary. For instance, LIME based approaches~\citep{ribeiro2016should, shrikumar2016not, bach2015pixel} locally approximate complex models with linear fits. Decision Trees are also considered more explainable if they are not too large. These approaches inherently assume a tradeoff between model performance and explanability, as less complex model classes tend to be empirically sub-par in performance relative to the success of the target black-box models they endeavor to explain. The \xgems\,  framework, however, does not rely on local approximations to provide explanations or assume such a trade-off. %Additionally, like complex models, these local approximations may themselves be prone to adversarial attacks, thus changing the explanations themselves. All example based explanation models, including those based on adversarial attacks, rely on data samples existing in the training set.  

We summarize our key contributions as follows:
\begin{inparaenum}
\item We introduce \xgems, a framework for explaining supervised black-box models via examples generated along the underlying data manifold.
\item We demonstrate the utility of \xgems \, ~in \begin{inparaenum} \item detecting bias in learned models, \item characterizing the probabilistic decision manifold w.r.t. examples, and \item facilitating model comparison beyond standard performance metrics.\end{inparaenum}
\end{inparaenum}

%We therefore leverage recent advances of characterizing such a data manifold of relevant training data using implicit generative models. By assuming access to such a generative model, we explain model behavior without relying on existing samples in the training dataset.
\section{Background}
\paragraph{Implicit Generative Models}can be described as stochastic procedures that generate samples (denoted by the random variable $\bx \in \mathbb{X}^d$) from the data distribution $p(\bx)$ without explicitly parameterizing $p(\bx)$. The two most significant types are the Variational Auto-Encoders (VAEs)~\citep{kingma2013auto} and Generative Adversarial Networks (GANs)~\citep{goodfellow2014generative}. Implicit generative models generally assume an underlying latent dimension $\bz \in \mathbb{R}^k$ that is mapped to the ambient data domain $\bx \in \mathbb{R}^d$ using a deterministic function $\mathcal{G}_{\theta}$ parametrized by $\theta$, usually as a deep neural network. The primary difference between GANs and VAEs is the training mechanism employed to learn function $\mathcal{G}_{\theta}$. GANs employ an adversarial framework by employing a discriminator that tries to classify generated samples from the deterministic function versus original samples and VAEs maximize an approximation to the data likelihood. The approximation thus obtained has an encoder-decoder structure of conventional autoencoders~\citep{doersch2016tutorial}. One can obtain a latent representation of any data sample within the latent embedding using the trained encoder network. While GANs do not train an associated encoder, recent advances in adversarially learned inference like BiGANs~\citep{dumoulin2016adversarially,donahue2016adversarial} can be utilized to obtain the latent embedding. In this work, we assume access to an implicit generative model that allows us to obtain the latent embedding of a data point.

Let $\mathcal{F}_{\psi} : \mathbb{R}^d \rightarrow \mathbb{R}^k$ (parametrized by $\psi$) be the inverse mapping function that provides the latent representation for a given data sample. Let $\mathcal{L} : \mathbb{R}^d \times \mathbb{R}^d \rightarrow \mathbb{R}_{+}$ be the analogous loss function such that for a given data sample $\tilde{\bx}$:
\begin{equation}
\tilde{\bz} = \argmin_{\bz} \mathcal{L}(\tilde{\bx}, \mathcal{G}_{\theta}(\bz)) \triangleq \mathcal{F}_{\psi}(\tilde{\bx})
\end{equation}
Examples of $\mathcal{F}_{\psi}$ are the encoder in a VAE, or an inference network in a BiGAN. An appropriate distance function in the data domain can be used as the loss $\mathcal{L}$.

Without loss of generality, we assume that we would like to provide explanations for a binary classifier. Let $y \in \{-1, 1\}$ be the target label. Let $f_{\phi} : \mathbb{R}^{d} \rightarrow \{-1, 1\}$ be the target black-box classifier to be `explained' and $\ell (f_{\phi}(\bx), y)$ be the loss function used to train the black-box classifier.

\paragraph{Adversarial Criticisms} Adversarial criticisms to explain black-box classifiers look for perturbations $\delta_{\bx}$ to data samples $\bx$ such that the perturbations maximize the loss $\ell (f_{\phi} (\bx + \delta_{\bx}), y)$ or change the predicted label. These perturbations are invisible to the human eye. That is, if $\tilde{\bx}$ is the target adversarial sample,  an adversarial attack solves a Taylor approximation to the following:
\begin{equation}\label{eq:adv}
\tilde{\bx} = {\argmax}_{\tilde{\bx} : \| \tilde{\bx} - \bx \|_p < \epsilon} \ell(f_{\phi} (\tilde{\bx}), y)
\end{equation}
%This leads to the following gradient steps to recover an adversarial criticism: \begin{inparaenum} \item[] $\bx \leftarrow \bx + \epsilon \cdot sign(\nabla_{\bx}\ell (\bx, y))$ when $p = \infty$ (where $sign(\cdot)$ refers to the sign operator), \item[] $\bx \leftarrow \bx + \epsilon \cdot \nabla_{\bx} \ell(\bx, y)$ when $p = 2$. \end{inparaenum}

We now characterize the proposed model and detail the kinds of explanations it can provide.

\section{Generating \Counterfactuals}
\begin{algorithm}[t]
\caption{{\bf{ Find $(\bx^*, y^*)$--\Counterfactual}} \\Input:  $(\bx^*, y^*) \in \mathbb{R}^{d} \times \{-1, 1\},  y_{tar}, \mathcal{G}_{\theta}, \mathcal{F}_{\psi}, f_{\phi}, \lambda, \eta > 0$}\label{alg:counterfactual}
\begin{algorithmic}
\STATE{Initialize $\bz = \mathcal{F}_{\psi}(\bx^*)$}
\WHILE{Not converged}
\STATE{$\tilde{\bz} \leftarrow \tilde{\bz} + \eta \nabla_{\tilde{\bz}}(\mathcal{L}(\bx^*, \mathcal{G}_{\theta}(\bz)) + \lambda \ell(f_{\phi}(\mathcal{G}(\bz)), y_{tar}))$}
\ENDWHILE
\STATE{$\tilde{\bx} = \mathcal{G}_{\theta} (\tilde{\bz})$}
\STATE{Return $\tilde{\bx}$}
\end{algorithmic}
\end{algorithm}
To provide explanations via examples over more \emph{naturalistic} perturbations, we introduce a new set of examples, called \emph{\counterfactuals} or \xgems. First, we train an implicit generative model $\mathcal{G}_{\theta}$ and an encoder network $\mathcal{F}_{\psi}$. %Note that we do not use a conditional generative model as training generative models is extremely computationally expensive and should not require retraining based on the target black-box task that we hope to analyze. Additionally, a shared manifold across different classification tasks allows to explain the model w.r.t. a common underlying manifold. 
\begin{equation}
\tilde{\bx} = \mathcal{G}_{\theta}({\argmin}_{ \bz \in \mathbb{R}^d} \mathcal{L}(\bx^*, \mathcal{G}_{\theta}(\bz)) + \lambda \ell(f_{\phi}(\mathcal{G}(\bz)), y_{tar}))
\label{eq:cost_fn}
\end{equation}
A \counterfactual \, is defined w.r.t. a given data sample $\bx^*$.

\begin{definition}[$\bx^*, y^*$-\Counterfactual]\label{def:counterfactual}
An {\bf{xGEM}} \, corresponding to a data point ($\bx^*, y^*$) and a target label $y_{tar} \neq y^*$, refers to the solution of Equation~\eqref{eq:cost_fn} for a fixed and known $\lambda>0$. The {\bf{xGEM}} is denoted by $\tilde{\bx}$.
\end{definition}

We propose Algorithm~\ref{alg:counterfactual} to estimate a \counterfactual \,{\bf{xGEM}} for any data point $\bx^*$. Intuitively, for a point $\bx^*$, we determine its latent representation using $\mathcal{F}_{\psi}$. This allows us to explain model behavior from a common latent representation across all black-boxes. To find realistic perturbations to this point, along the data manifold, we traverse the latent space of the generator $\mathcal{G}_{\theta}$ (our proxy for the data manifold) until the label switches to the desired target label $y_{tar}$. The desired \emph{\counterfactual} or {\bf{xGEM}} is the sample generated at the switch point in the latent embedding. %and can be easily recovered using the proxy generator $\mathcal{G}_{\theta}$.

We empirically highlight the benefits of the discovering \counterfactuals\,  in different contexts and abstractions that provide insights into model behavior.

\section{Explanations using \Counterfactuals}
We first use a simple setting with simulated data to highlight the differences between the proposed explanation tool compared to criticisms and prototypes derived from adversarial attacks~\citep{stock2017convnets}.
\subsection{An alternative view to Adversarial Criticisms}
\begin{figure}[!htbp]
  \begin{subfigure}{.55\linewidth}
   \centering\includegraphics[width=.35\linewidth]{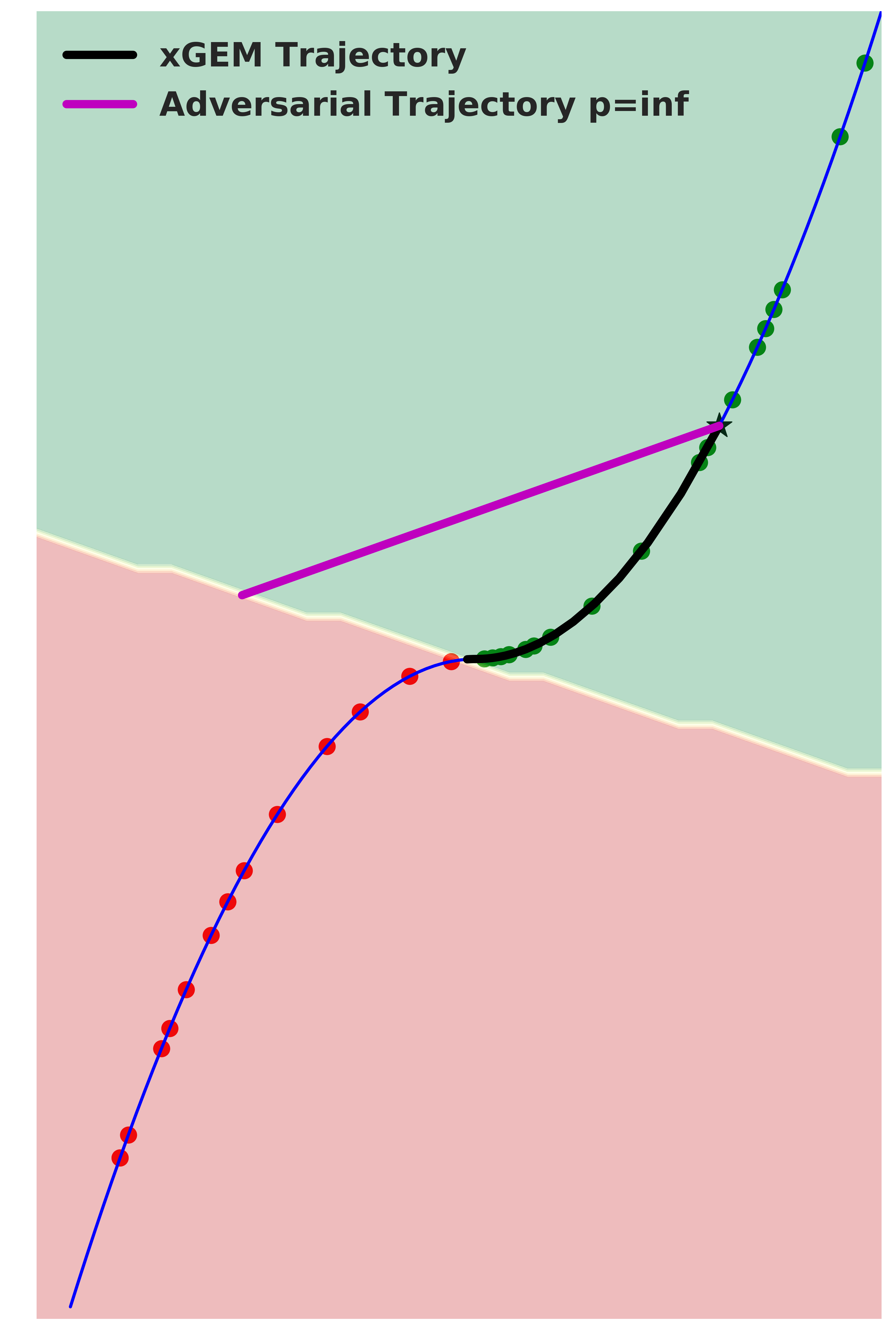}
   \includegraphics[width=.35\linewidth]{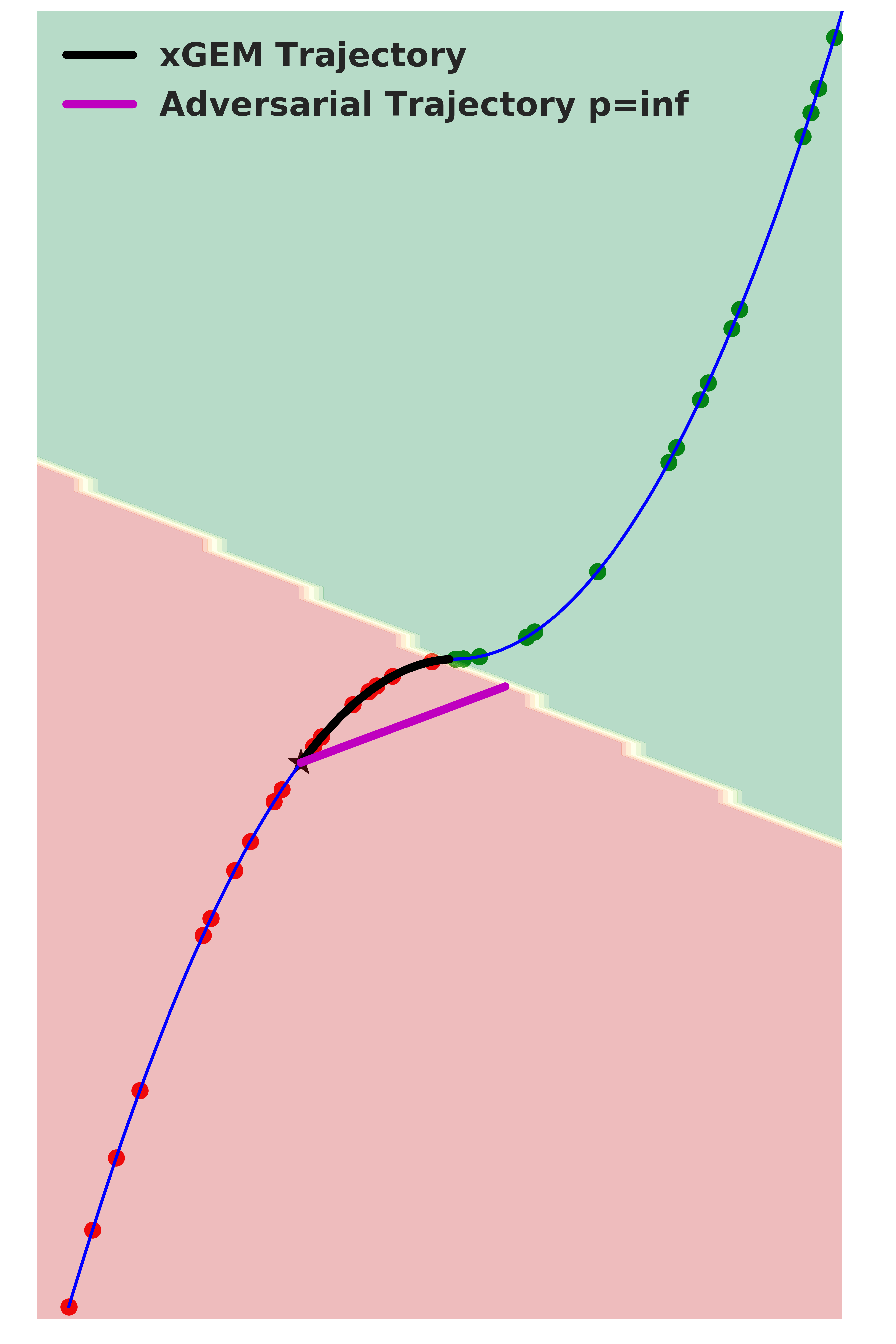}
    \caption{}
    \vspace{-15px}
    \label{fig:simulated_data}
  \end{subfigure}%
  \captionsetup[subfigure]{labelformat=empty}
  \begin{subfigure}{.42\linewidth}
     \centering
\caption{\emph{\Counterfactuals} versus \emph{Adversarial} criticisms~\citep{stock2017convnets}, for a parabolic manifold (shown in blue). Green points belong to class 1 and red points to class 2. The black trajectories in all figures are gradient steps taken by Algorithm~\ref{alg:counterfactual} while the magenta trajectories correspond to adversarial trajectories determined by Equation~\ref{eq:adv} with $p = \infty $. Note that all decision boundaries in Figures (a) and (b) separate the data. The decision boundary is trained by optimizing a softmax regression using the cross-entropy loss function.}
  \end{subfigure}
  \caption{}
\end{figure}

Figure~\ref{fig:simulated_data} demonstrates a linear decision boundary trained on data with ambient dimension equal to 2. The one-dimensional data manifold is parabolic as shown by the blue curve. The green points are in class 1 and red points are samples belonging to class 2. The figure illustrates \counterfactuals \, as well as the trajectory taken by the gradient steps of Algorithm~\ref{alg:counterfactual}. The trajectory to generate an adversarial criticism stems from Equation~\eqref{eq:adv}. A generative model maps from a 1d latent dimension to the data manifold shown by the blue curve. A single layer (softmax) neural network with output dimension=2 is trained on points sampled from this manifold (the yellow decision boundary separates the two classes -- regions marked by the pink and green regions). As demonstrated by the figure, navigating along the latent dimension of the generator encourages the {\bf{xGEM}} trajectory to be constrained along the data manifold, while adversarial criticisms may lie well outside the manifold. Thus \emph{\counterfactuals} offer alternative view of classifier behavior via examples. 

We defer examples of {\bf{xGEMs}} evaluated for the MNIST dataset to the Appendix in the interest of space.
\subsection{Towards automated bias detection}

We now demonstrate the utility of generating \counterfactuals \, to detect if a target classifier is \confounded w.r.t. a given attribute of interest. In particular, we wish to determine whether a black-box is differentiating among the target labels using spurious correlations in the data. For instance, a classifier trained to determine the best medical intervention may be relying on attributes like gender to determine best treatment. It is desirable to have an automated mechanism to detect such behavior. We say that a classifier is \confounded with an attribute of interest $a$ if the attribute $a$ substantially influences the black-box's predictions. We make this concrete in the context of our framework below.

\begin{minipage}{\textwidth}
\begin{minipage}{0.53\textwidth}
\centering
\begin{figure}[H]
\includegraphics[width=0.85\textwidth]{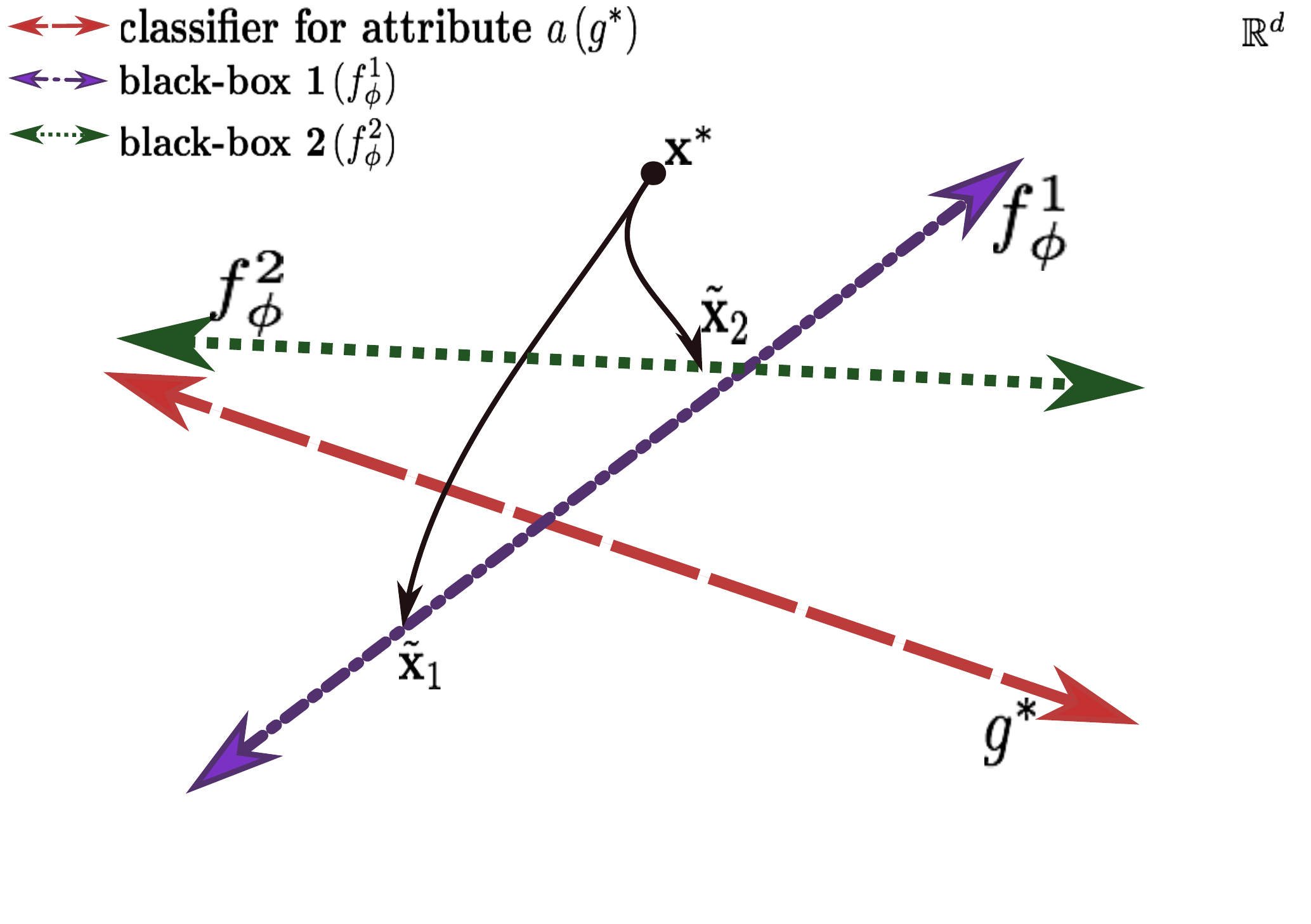}
\caption{Example of bias detection. Target black-boxes:$f^1_{\phi}$ and $f^2_{\phi}$. $g^*$ classifies points w.r.t. $a$. $\tilde{\bx}_1$ and $\tilde{\bx}_2$ are {\bf{xGEMs}} corresponding to $\bx^*$ for $f^1_{\phi}$ and $f^2_{\phi}$ resp. $\tilde{\bx}_2$'s attribute prediction (w.r.t $g^*$) is the same as that of $\bx^*$ while that of $\tilde{\bx}_2$ is different. Thus we say that $f^1_{\phi}$ is biased w.r.t. attribute $a$ for sample $\bx^*$.}
\label{fig:confounding}
\end{figure}
\end{minipage}
\hspace{2px}
\begin{minipage}{0.3\textwidth}
\begin{table}[H]
\centering
%\vspace{-50px}
\begin{tabular}{p{2.2cm}ll}
\toprule
& \multicolumn{2}{c}{Target black-box label} \\
\cmidrule(r){2-3}
Attribute ($a$) Classifier     & Black Hair     & Blond Hair \\
\midrule
$\hat{g}$ (orig) & FP:0.003  & FP:0.000     \\
& FN:0.002  & FN:0.018     \\
& Acc: 0.997  & Acc:0.999    \\
\midrule
$\hat{g}$ (recalibrated) & FP:0.003  & FP:0.003 \\
& FN:0.018  & FN:0.018  \\
& Acc:0.989  & Acc:0.996 \\
\bottomrule
\end{tabular}
\vspace{5px}
\captionof{table}{Recalibrated Gender Classifier.}
\label{tab:eq_odds_details}
\end{table}
\end{minipage}
\end{minipage}

Without loss of generality let $a \in\{-1, 1\}$ be the (potentially protected) binary attribute of interest. We wish to examine whether the target classifier $f_{\phi}$ is biased/\confounded by $a$. Intuitively, we hope that attribute $a$ of an {\bf{xGEM}} should be the same as that of the original point. In order to detect this, we assume there exists an oracle $g^* : \mathbb{R}^d \rightarrow \{-1, 1\}$ that perfectly classifies  the \confounding attribute $a$ when considered as the dependent variable, based on the other ($d$) independent variables. Additionally, we assume that $g^*$ is not \confounded by the target label of the black-box $y$ and is not used by $g^*$ to predict $a$. Let $\mathbb{R}^d \times \{-1, 1\} \times \{-1, 1\} \supset \mathcal{D} \triangleq \{(\bx_i, y_i, a_i) \, , i \in [N] \}$ be the training data where $i$ indexes a given point. Let $\tilde{\bx}_i$ be the {\bf{xGEM}} of $\bx_i$ w.r.t. $f_{\phi}$ as returned by Algorithm~\ref{alg:counterfactual}. We argue that classifier $f_{\phi}$ is \confounded by the attribute $a$ if equation~\eqref{eq:conf} holds for a given $\delta >0$.
\begin{equation}\label{eq:conf}
\frac{E_{\mathcal{D}}[\mathbbm{1}(g^*(\tilde{\bx}) \neq a)]}{|\mathcal{D}|} > \delta
\end{equation}

In practice, access to a perfect oracle $g^*$ is infeasible or prohibitively expensive. In some cases, such a classifier may be provided by regulatory bodies, thereby adhering to predetermined criterion as to what accounts for a \emph{reliable} proxy oracle. For this case study, we assume it is sufficient that the proxy oracle has the same false positive and false negative error rates w.r.t. the target label, which is a fairness condition known as the Equalized Odds Criterion~\citep{hardt2016equality}. To demonstrate our algorithm, we assume access to a proxy oracle $\hat{g} : \mathbb{R}^d \rightarrow \{-1, 1\}$ that satisfies the following conditions, given a $0.5 \ll \tau < 1$: \\\begin{inparaenum}\item[(i) ]\label{cond:cond1} $E_{\mathcal{D}}[\mathbbm{1}(\hat{g}) == a) ] > \tau$ \item[(ii)] \label{cond:cond2} $\hat{g}$ satisfies the Equalized Odds~\citep{hardt2016equality} criterion w.r.t. the target label $y$. \end{inparaenum}

\begin{minipage}{\textwidth}
  \begin{minipage}[b]{0.5\textwidth}
    \centering
      \begin{table}[H]
       \begin{tabular}{p{1.4cm} ll}
       \toprule
    Black-box Classifier & Accuracy & \Confounding \, metric\\
       \midrule
       $f_{\phi}^1$ & 0.9933 & 0.1704 \\
 & & \\
       $f_{\phi}^2$ &0.9155& 0.4323 \\
       \bottomrule
       \end{tabular}
       \vspace{16px}
       \captionof{table}{\Confounding metric}
      \label{tab:class_details}
       \end{table}
  \end{minipage}%
  \hspace{2px}
  \begin{minipage}[b]{.5\textwidth}
    \centering
    \begin{table}[H]
    \centering
  \begin{tabular}{lll}
    \toprule &
   \multicolumn{2}{c}{Target label} \\
    \cmidrule(r){2-3}
    Black-box & Black Hair     & Blond Hair \\
    \midrule
    $f_{\phi}^1$ & Male:0.4550  & Male:0.1432     \\
                 & Female:0.0159  & Female:0.0484     \\
                 & Overall:0.2430  & Overall:0.0539     \\
     \midrule
    $f_{\phi}^2$ & Male:0.7716  & Male:0.1475 \\
                 & Female:0.0045  & Female:0.5024  \\
                 & Overall:0.4012  & Overall:0.4821  \\
    \bottomrule
  \end{tabular}
  \vspace{2px}
   \captionof{table}{\Confounding metric by gender}
      \label{tab:strat_details}
       \end{table}
    \end{minipage}
  \end{minipage}

Note that while we consider $\hat{g}$ as an inexpensive proxy for $g^*$, we prescribe that the experiment be carried out with $g^*$. Figure~\ref{fig:confounding} demonstrates how such \confounding could be detected, as well as used for model comparison w.r.t. their biases. As shown in the figure, $f_{\phi}^1$ and $f_{\phi}^2$ are the classification boundaries of two black-box models classifying a target label of interest. $g^*$ is a classifier that classifies the data according to attribute $a$. Consider the sample $\bx^*$ and let $\tilde{\bx}_1$ and $\tilde{\bx}_2$ be the \counterfactuals \, of $\bx^*$ corresponding to classifiers $f_{\phi}^1$ and $f_{\phi}^2$ respectively. As shown in the figure, the attribute $a$ of the {\bf{xGEM}} $\tilde{\bx}_1$ is different from that of $\bx^*$ while that of $\tilde{\bx_2}$ is not. We conclude that a black-box $f^1_{\phi}$ is \confounded if the fraction of points whose \counterfactuals \, or {\bf{xGEM}}s \, that change attribute $a$ is greater than $\delta$. Thus an empirical estimate of Equation~\eqref{eq:conf} gives a metric that can quantify the amount of \confounding in a given black-box, while also allowing to compare different black-boxes w.r.t. the target attribute $a$.

We evaluate our framework for \confounding detection in facial images using the CelebA~\citep{liu2015faceattributes} dataset. The target black-box classifier predicts the binary facial attribute -- hair color (black or blond). We determine whether or not the black-box is \confounded with gender. We restrict to two genders, male and female, based on annotations available in CelebA. In particular, $\hat{g}$ is a ResNet model~\citep{he2016deep}\footnote{\url{https://github.com/tensorflow/models/tree/master/tutorials/image/cifar10_estimator}} that classifies celebA faces by gender . $\hat{g}$ is recalibrated to satisfy the two conditions mentioned earlier. 
%~\ref{cond:cond1} and~\ref{cond:cond2}. 
Details of $\hat{g}$'s performance and recalibration are provided in Table~\ref{tab:eq_odds_details}. 

Two ResNet models $f_{\phi}^1$ and $f_{\phi}^2$ are trained to detect the hair color attribute (black hair vs blond hair) using two different datasets. $f_{\phi}^1$ is trained on all face samples with either black or blond hair whereas $f_{\phi}^2$ is trained such that all black hair samples are male while blond haired samples are all female. Table~\ref{tab:class_details} gives the overall validation accuracy of both classifiers. Note that the validation set used for $f_{\phi}^1$ and $f_{\phi}^2$ are the same. 

\begin{figure}[!ht]
\setlength{\floatsep}{2pt plus 1.0pt minus 2.0pt}
  \includegraphics[width=\textwidth]{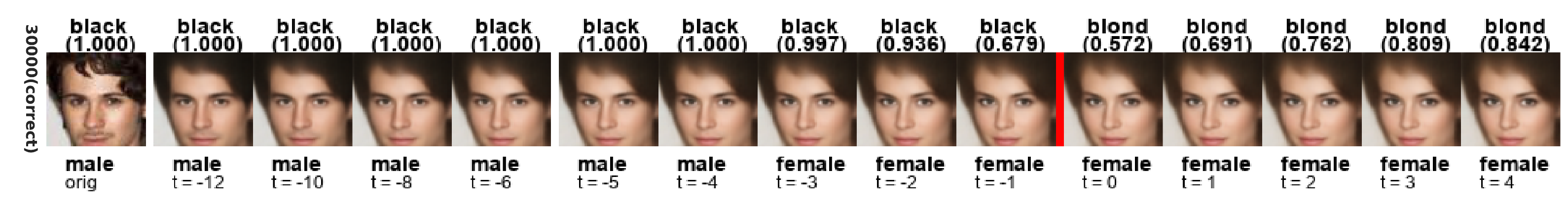}
      \vskip -12px
      \includegraphics[width=\textwidth]{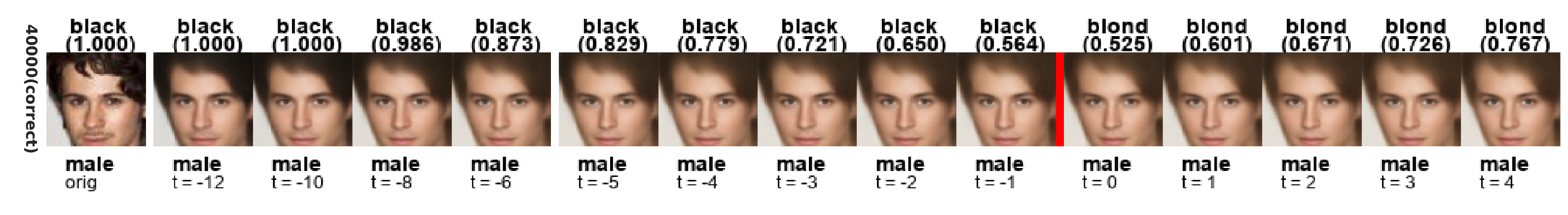}
      \medskip
     \includegraphics[width=\textwidth]{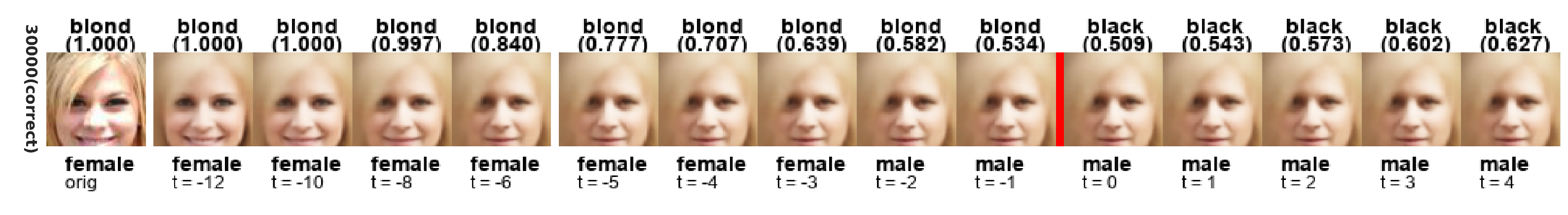}
     \vskip -18px
     \includegraphics[width=\textwidth]{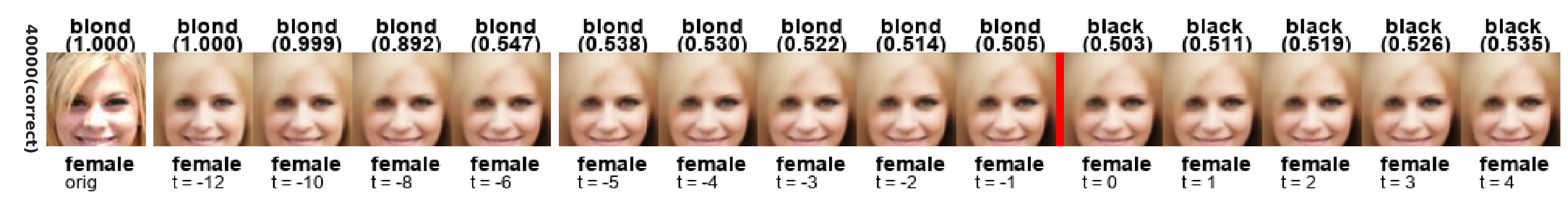}
  \caption{We test whether ResNet models $f^1_{\phi}$ and $f^2_{\phi}$, both trained to detect hair color but on different data distributions are \confounded with gender. Two samples for classifiers $f^1_{\phi}$ (first sub row) and $f^2_{\phi}$ (second sub row) are shown. The leftmost image is the original figure, followed by its reconstruction from the encoder $F_{\psi}$. Reconstructions are plotted as  Algorithm~\ref{alg:counterfactual} (with $\lambda=0.01$) progresses toward crossing the decision boundary. The red bar indicates change in hair color label indicated at the top of each image along with the confidence of prediction. The label at the bottom indicates gender as predicted by $\hat{g}$. For both samples, classifier $f^1_{\phi}$, trained on biased data changes the gender ($1^{st}$ and $3^{rd}$ rows) while crossing the decision boundary whereas the other black-box does not.}
  \label{fig:confounding_faces}
\end{figure}

Table~\ref{tab:class_details} also shows the fraction of samples whose \counterfactuals'  predicted attribute $a$ (in this case gender) is different from the original training sample w.r.t. $\hat{g}$. The fraction of \confounded \, samples is clearly much larger for the classifier trained on a biased dataset as determined by the proxy oracle $\hat{g}$. Additionally, Table~\ref{tab:strat_details} suggests a 10--fold increase in the fraction of \confounding for blond haired females with the biased classifier $f_{\phi}^2$. Notice the decrease in the amount of \confounding for black haired females while a general increase in \confounding for all black haired faces. As an aside, the biased model $f_{\phi}^2$ also changes the background more than hair color in order to change the hair color label (see Figure~\ref{fig:confounding_faces}). This suggests that quantifying such \confounding using \counterfactuals allows us to characterize biases w.r.t. any attribute of interest.

Figure~\ref{fig:confounding_faces} shows a few examples of such \confounded \, images for the two black-boxes. In particular, we show examples where the black-box trained on biased data for hair color classification changes gender of the sample as it crosses the decision boundary whereas the black-box trained on unbiased data does not\footnote{All qualitative figures were chosen based on the confidence of the prediction from the black-box and confidence of the reconstructed image}.

\subsection{Case Study: Model Assessment beyond performance metrics}
\begin{figure}[!ht]
%\captionsetup[subfigure]{justification=justified,singlelinecheck=false}
 %\begin{subfigure}{.5\linewidth}
 \includegraphics[width=\linewidth]{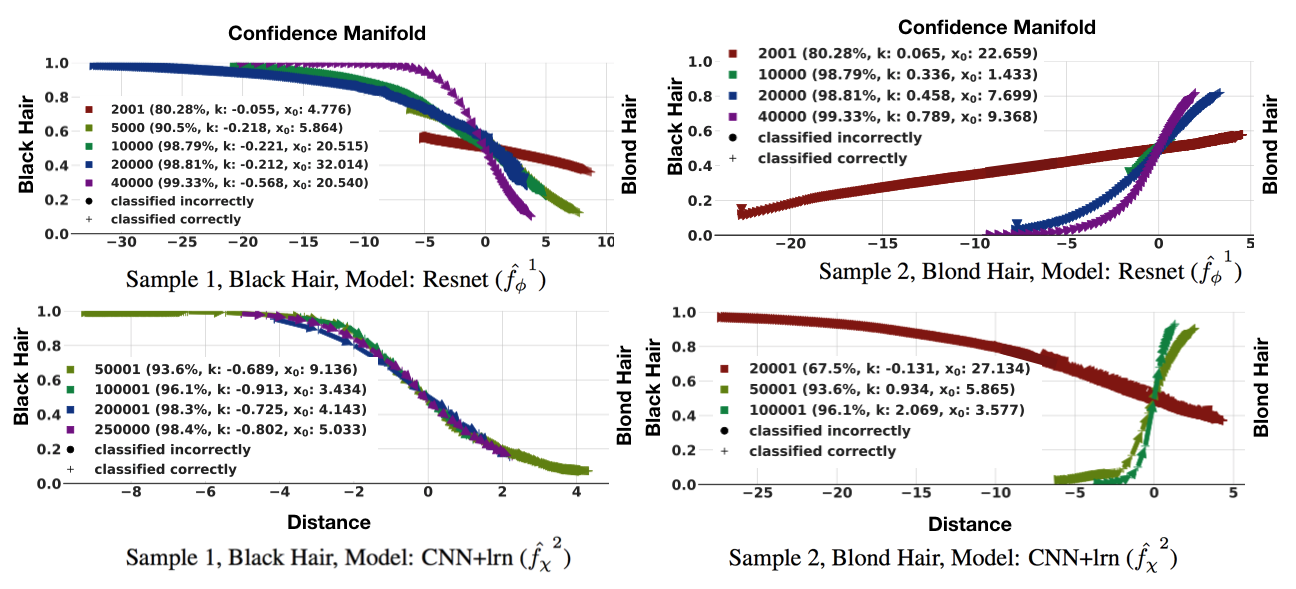}
    \caption{Confidence manifolds for a few data samples for black-box models 1 and 2.}
  %  \includegraphics[width=\linewidth]{figures/training/cnn_female_black_manifold_33.pdf}
%    \caption{Sample 1, Black Hair, Model: CNN+lrn ($\hat{f_{\chi}}^2$)}
   % \end{subfigure}%
   % \begin{subfigure}{.5\linewidth}
    %    \includegraphics[width=\linewidth]{figures/training/resnet_female_blond_manifold_170.pdf}
 %   \caption{Sample 2, Blond Hair, Model: Resnet  ($\hat{f_{\phi}}^1$)}
    % \includegraphics[width=\linewidth]{figures/training/cnn_female_blond_manifold_170.pdf}
 %    \caption{Sample 2, Blond Hair, Model: CNN+lrn ($\hat{f_{\chi}}^2$)}
    %\end{subfigure}
    %\caption{Confidence manifolds for a few data samples for black-box models 1 and 2.}
\label{fig:training_mani}
  \end{figure}
    
An important aspect of black-box analyses is to study the progression of training complex models. Specifically, observing \counterfactuals\,  allows us to consider model behavior in the following aspects: \begin{inparaenum}
\item[1)] Discerning shifts in features relied on by the black-box to differentiate between classes during training.
\item[2)] Characterizing the probabilistic manifolds of \counterfactuals\, as training progresses and its relation to calibration of complex networks~\citep{degroot1983comparison}.
\item[3)] Qualitative trade-offs and/or mistakes made by the classifier for prototypical examples.
\end{inparaenum}

Reliability Diagrams have been used as a summary statistic to evaluate model calibration~\citep{degroot1983comparison} that aims to study whether the confidence of a prediction matches the ground truth likelihood of the prediction. It has been observed that while model performance has improved substantially in recent years because of deep networks, such models are typically more prone to mis-calibration~\citep{guo2017calibration}. We  provide a complementary statistic to Reliability Diagrams to assist model assessment/comparison. 

For this study we train two deep networks $\hat{f_{\phi}}^1$ (a ResNet model) and $\hat{f_{\chi}}^2$ (a four layer CNN with local response normalization (lrn)~\footnote{https://github.com/tensorflow/models/tree/master/tutorials/image/cifar10}) with CelebA face images for the hair color (black/blond) binary classification task. For a given face, we evaluate the corresponding {\bf{xGEM}} at multiple incremental training steps. We plot the confidence of labeling a point to have black hair with respect to the distance of the original reconstruction and its {\bf{xGEM}} including all intermediate points from the decision boundary (called `confidence manifold'). Thus, all samples originally labeled black should have high confidence of being labeled and the confidence decreases as the sample crosses the decision boundary (vice-versa for blond haired faces). Figure~\ref{fig:training_mani} shows the confidence manifolds for two samples (one in each column). 

The top and bottom rows represent the manifolds obtained during training for model 1 ($\hat{f_{\phi}}^1$) and  model 2 ($\hat{f_{\chi}}^2$) respectively. Sample 1(column 1) is a face with black hair while Sample 2(column 2) has blond hair. Legends show the distance of reconstructions from the original sample along the gradient steps, followed by overall classifier performance. Additionally, we fit a logistic function $f(x) = \frac{1}{1+\exp^{-k(x-x_0)}}$ to each curve. All plots have been shift-aligned using $x_0$. The confidence manifold for the same instance is fairly different across each model. As expected, the overall steepness increases as model trains to better discriminate samples. Intuitively, higher $x_0$ suggests that the classifier can easily discriminate the label with high confidence. For instance, for comparable overall accuracies, the manifolds suggest that model 2 has trained a decision boundary such that a \counterfactual \, is relatively close in image distance (compared to that of model 1). In the case of Sample 2, it is clear that model 2 mis-labels the data point with high confidence initially while learning to predict the correct label eventually. However, a decrease in $x_0$ as training progresses for both models suggests a significant shift of the decision boundary to be closer to Sample 2. Qualitative images corresponding to these manifolds are shown in the Appendix.
\begin{figure}[!ht]
 \begin{subfigure}{.5\textwidth}
 \centering
 \includegraphics[width=0.8\textwidth]{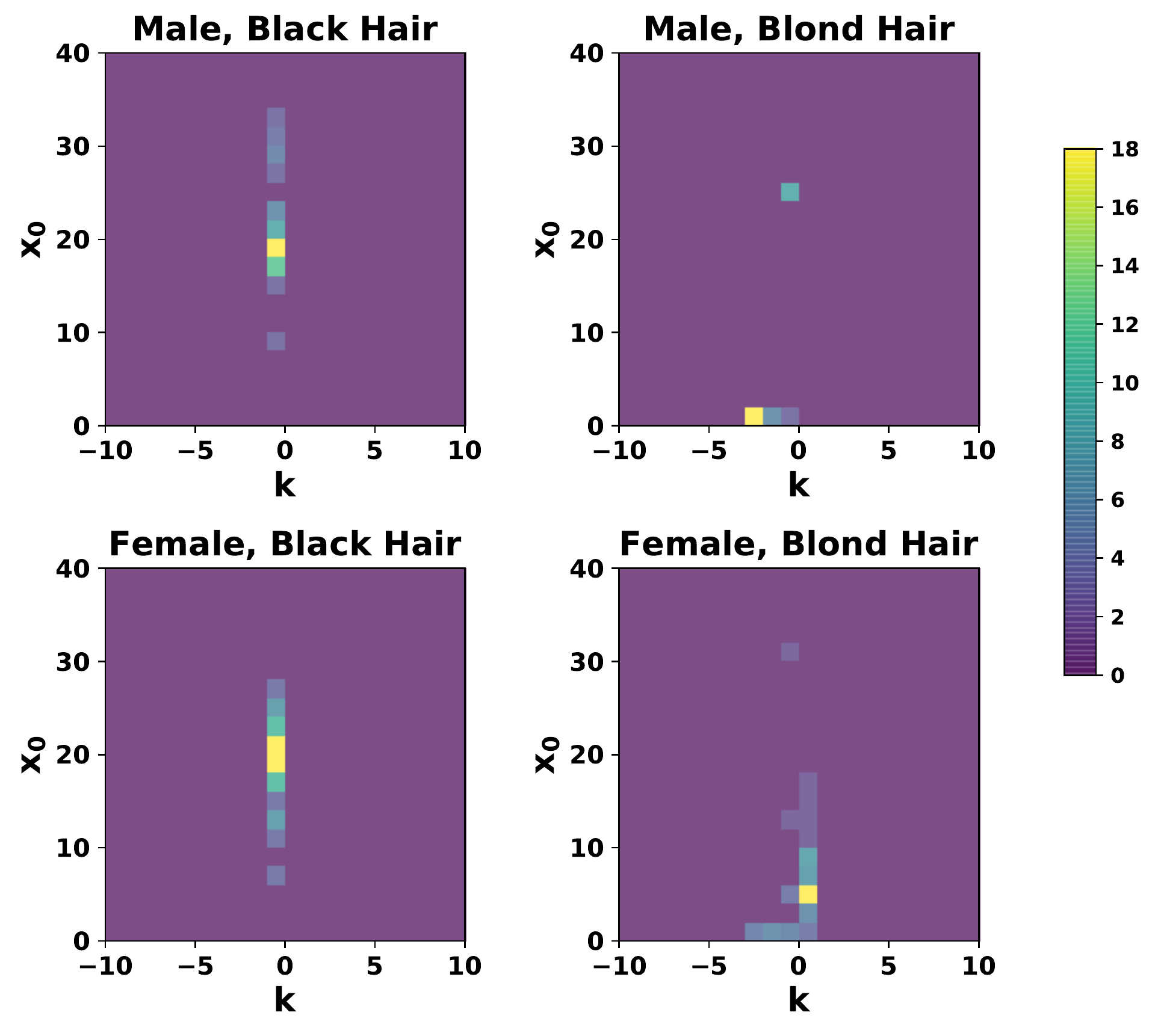}
 \caption{Black-box 1, ResNet ($\hat{f}_{\phi}^1$)}
 \label{fig:model_compare1}
 \end{subfigure}
   \hspace{-10px}
 \begin{subfigure}{.5\textwidth}
 \centering
 \includegraphics[width=0.8\textwidth]{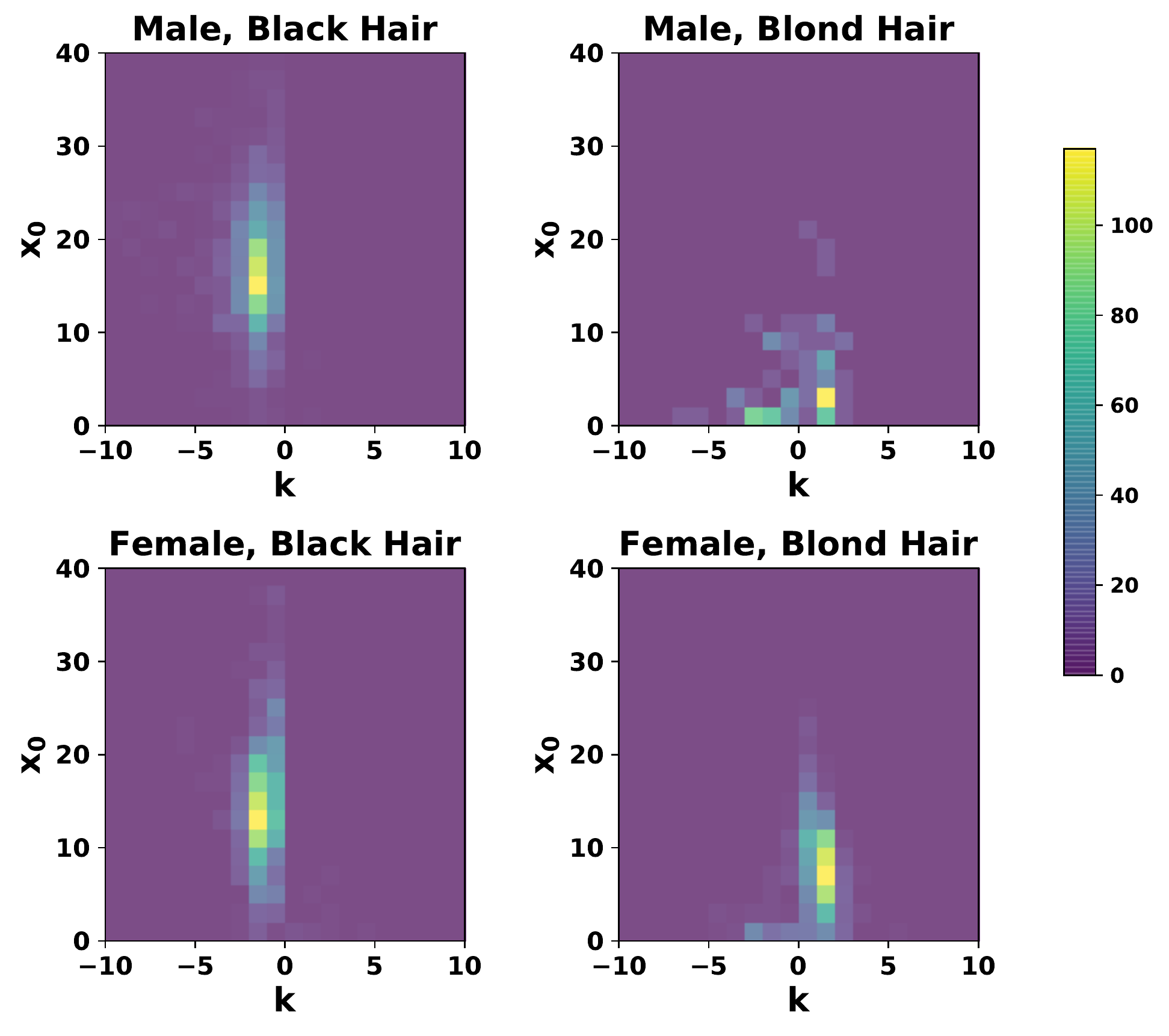}
    \caption{Black-box 2, CNN+lrn ($\hat{f}_{\chi}^2$)}
    \label{fig:model_compare2}
    \end{subfigure}
%     \hspace{-10px}
%     \begin{subfigure}{.3\textwidth}
%      \centering
%      \includegraphics[width=\linewidth]{figures/model_comparison/calibration_compare_last_iter.pdf}
% \caption{}
% \label{fig:model_compare3}
%   \end{subfigure}
  \caption{(a) and (b): 2d-Histograms of the parameters of the logistic function fits to the confidence manifolds for a $\sim4000$ samples.}% (c) Reliability Diagram for Calibration stratified by (potentially protected) attributes of interest (gender): A perfectly calibrated classifier should manifest an identity function. Deviation from the identity function suggests mis-calibration and can be used for model comparison when accuracy and other metrics are comparable. }
\label{fig:model_compare}
  \end{figure}
  
\begin{figure}[!ht]
\centering
\includegraphics[width=0.4\linewidth]{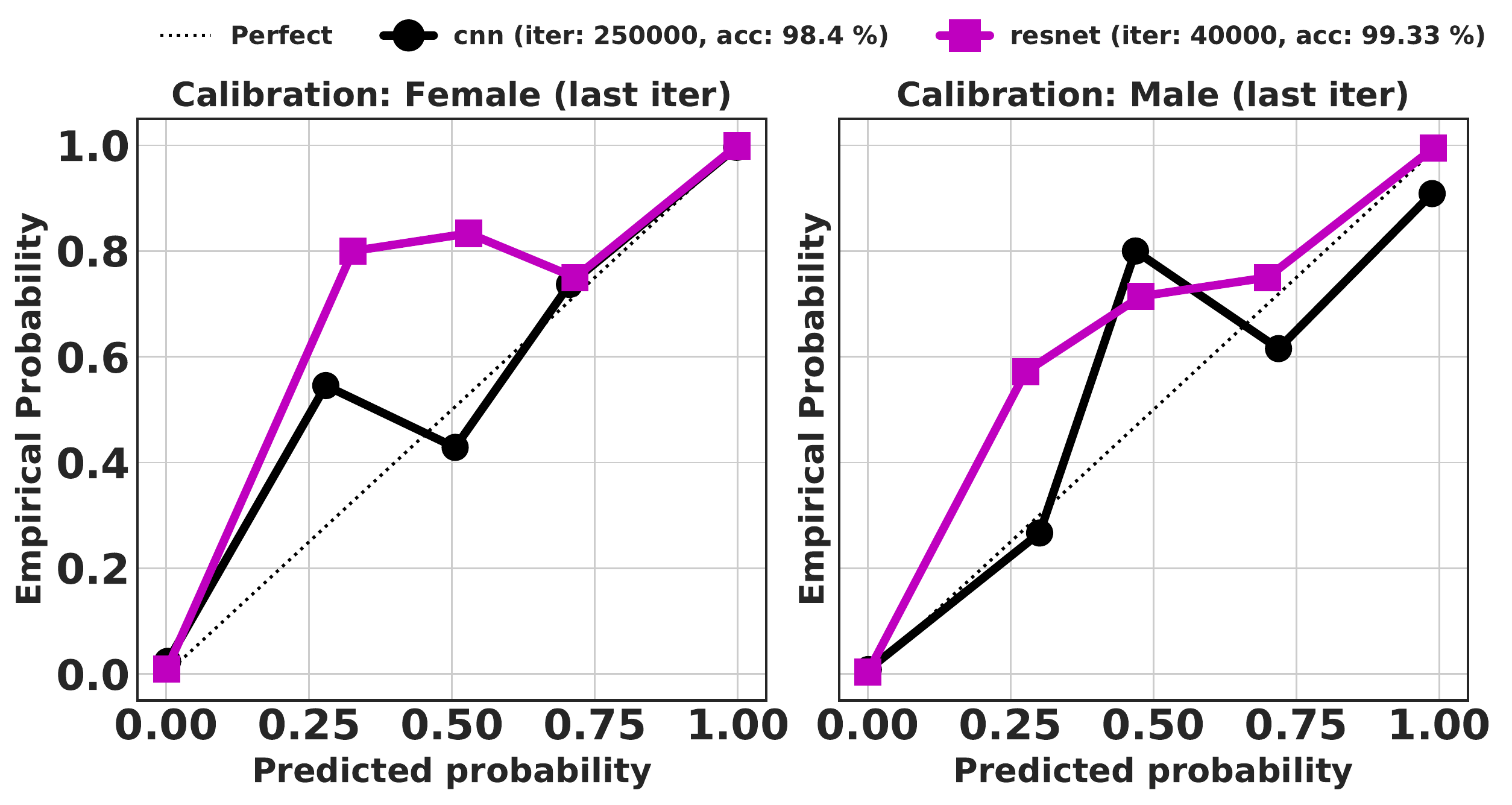}
\caption{Reliability Diagram for Calibration stratified by (potentially protected) attributes of interest (gender): A perfectly calibrated classifier should manifest an identity function. Deviation from the identity function suggests mis-calibration and can be used for model comparison when accuracy and other metrics are comparable.}
\label{fig:model_compare3}
\end{figure}
Figures~\ref{fig:model_compare1} and~\ref{fig:model_compare2} show the 2d histogram of the logistic function parameter estimates stratified by the target label and the attribute of interest (gender). This allows to summarize the confidence manifolds across groups of interest for overall model comparison. For reference, Figure~\ref{fig:model_compare3} shows the Reliability Diagram for both black-boxes. The ResNet model generally demonstrates more uniform steepness across samples at different distances from the decision boundary compared to the CNN+lrn model. Both models have a relatively small $x_0$ for blond haired males suggesting lower confidence in their predictions. Thus, summarizing confidence manifolds provides additional insight that may not be characterized by Reliability Diagrams for model comparison.

\section{Discussion}
This work presents a novel approach to characterizing and explaining black-box supervised models via examples. An unsupervised implicit generative model is used as to approximate the data manifold, and subsequently used to guide the generation of increasingly confounding examples given a starting point. These examples are used to probe the target black-box in several ways. In particular, we demonstrate the utility of \counterfactuals\, in automatically detecting bias in black-box learning w.r.t. a (potentially protected) attribute as well as for model comparison. The proposed method also allows one to visualize training progression and provides insights complementary to notions of calibration of the black-box model. Limitations of the proposed method include reliance on the implicit generator as a proxy of the data manifold. However, we note that we do not rely on specific architectures and/or training mechanisms for the generative model. We used images as they are easy to visualize even in high-dimensions. However extending our studies to complex datasets beyond images is a compelling future extension.

%\section*{References}
%\renewcommand\refname{}
\small
\bibliography{ttlv}
\bibliographystyle{plainnat}

%\subsubsection*{Acknowledgments}
\section*{Appendix}

\subsection*{{\bf{xGEMs}} for MNIST}
\begin{figure}[!htp]
%\centering
\includegraphics[width=0.8\linewidth]{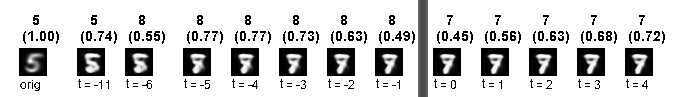}
\includegraphics[width=\linewidth]{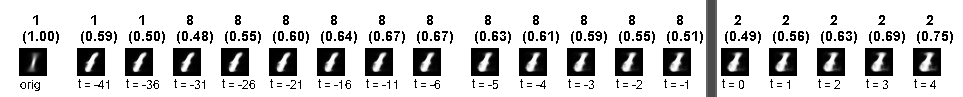}
\includegraphics[width=0.9\linewidth]{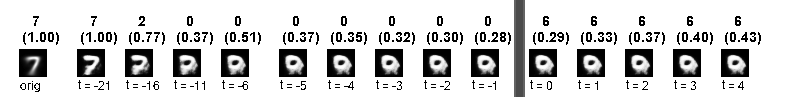}
\caption{\Counterfactuals \, for MNIST data. $\mathcal{G}_{\theta} : \mathbb{R}^{100} \rightarrow \mathbb{R}^{28\times28}$ is a VAE while the target black-box is a softmax classifier. Each row shows a \counterfactual transition for a single digit (labeled `orig'). The gray vertical bars indicate transition to the target label $\by_{tar}$. Reconstructions in each row are intermediate reconstructions obtained using Algorithm~\ref{alg:counterfactual}. The confidence of the clas prediction is shown in parentheses for each reconstruction.}
\label{fig:mnist_counterfactuals}
\end{figure}

Figure~\ref{fig:mnist_counterfactuals} shows \counterfactual s generated for a (multi-class) softmax classifier for MNIST\footnote{http://yann.lecun.com/exdb/mnist/} digit data. The first row in Figure~\ref{fig:mnist_counterfactuals} shows \counterfactual for digit $5$ if $\by_{tar} = 7$, while second and third row show \counterfactuals for digits $1$ and $7$ with $\by_{tar} = 2$ and $\by_{tar} = 6$ respectively. Notice how while traversing the manifold, the classifier switches decision from $5$ to $8$ and then to the target label $7$ (row 1). While the intermediate samples look like $7$ to human eye, the classifier is biased toward predicting $8$. Row 2 suggests a bias toward predicting $1$ as $8$ for a minor smudging (visible to human eye). Finally, the third row demonstrates how the \counterfactual for $7$ suggests that the classifier considers a $0$ to be labeled as $6$. Thus \counterfactuals \, can provide insight into the decision boundary of the classifier for each pair of digits.

\begin{figure}[!htpb]
\centering
\includegraphics[width=0.9\linewidth]{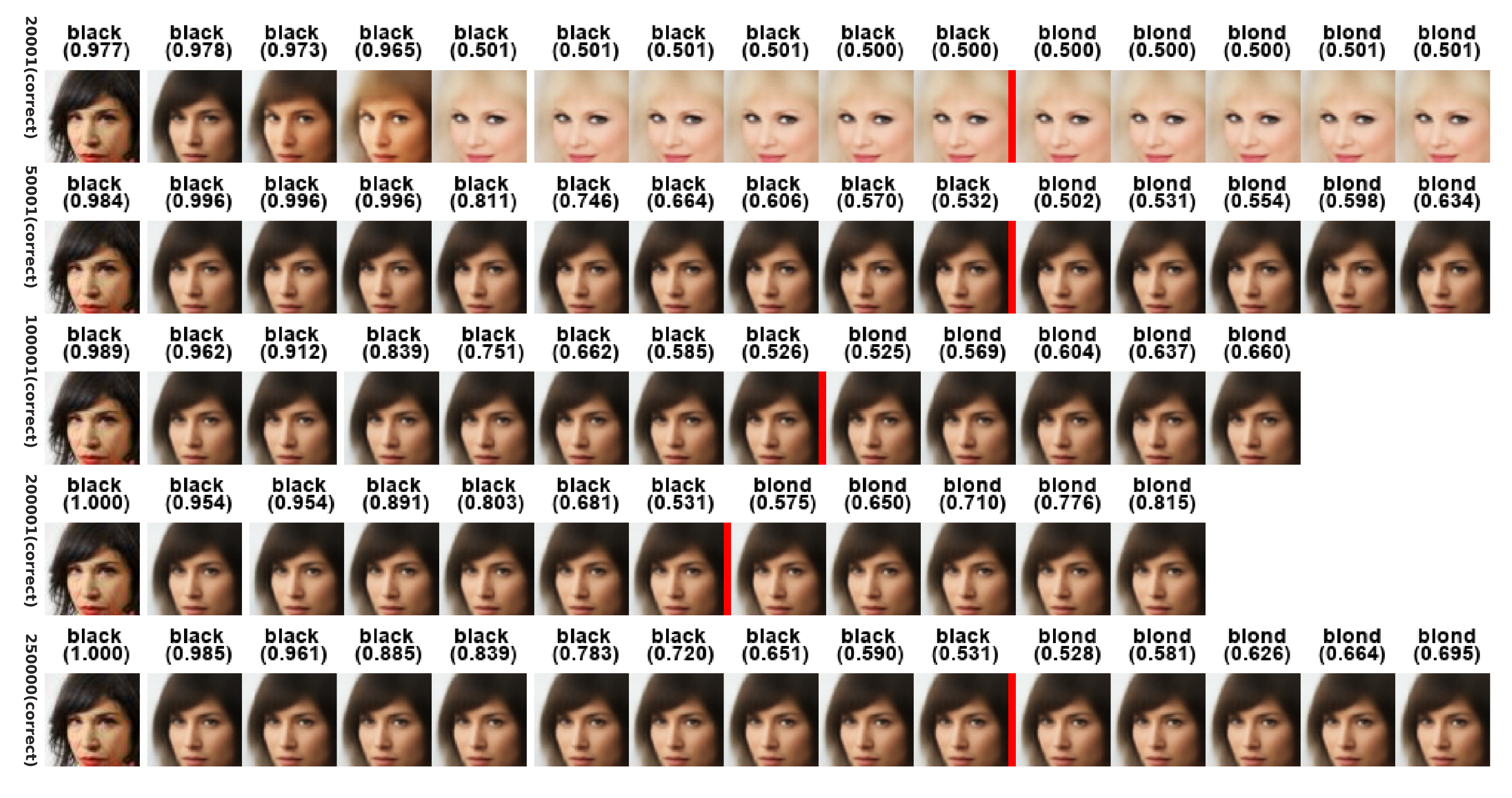}
\caption{Training progression for celebA face image for the CNN+lrn model.}
\label{fig:training_img_cnn}
\end{figure}

\subsection*{Case Study: Evaluating Model Training Progression}
\begin{figure}[t]
\centering
\includegraphics[width=0.9\linewidth]{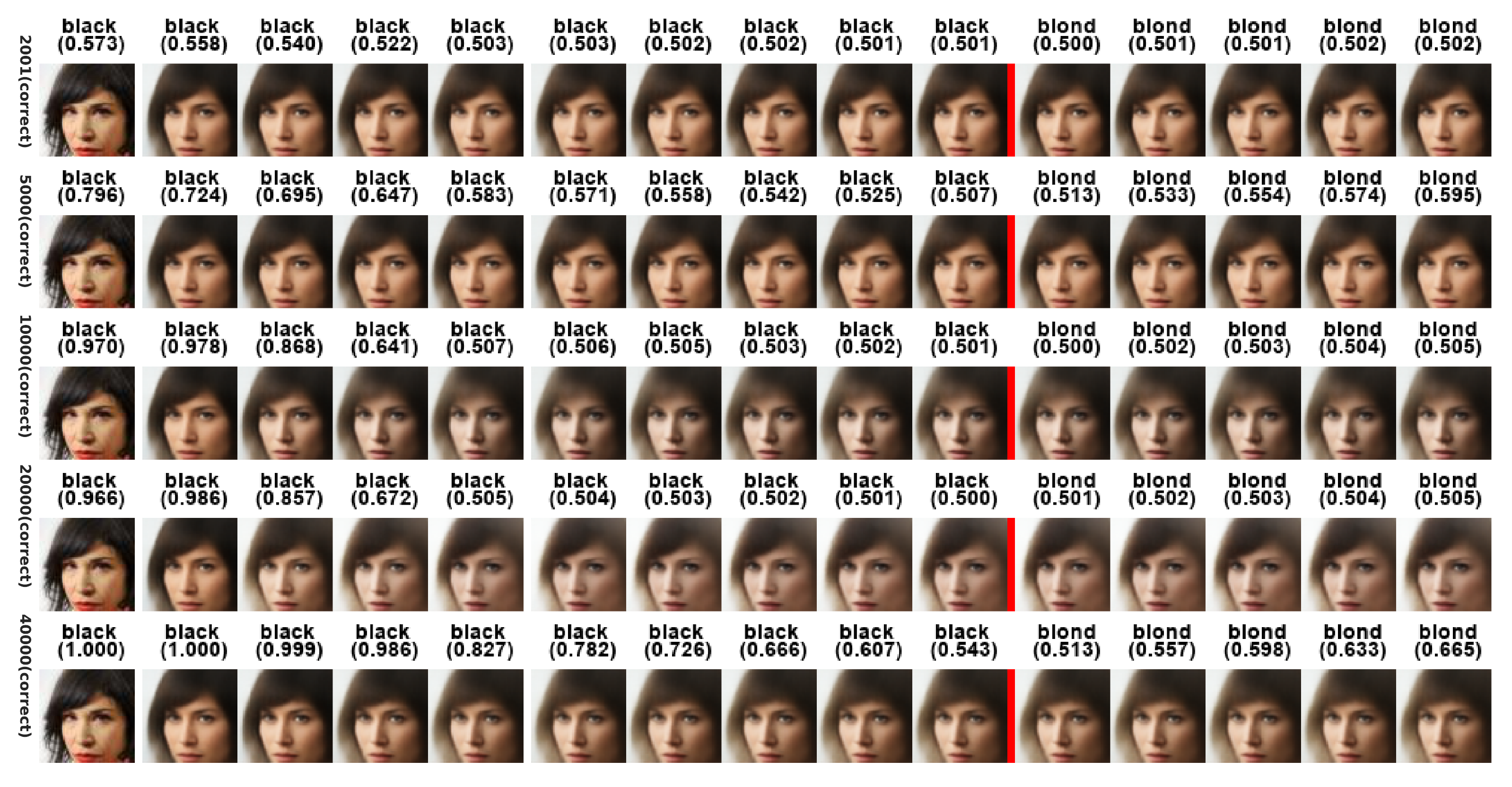}
\caption{Training progression for celebA face image for the ResNet model.}
\label{fig:training_img_resnet}
\end{figure}

Figures~\ref{fig:training_img_cnn} and~\ref{fig:training_img_resnet} show {\bf{xGEMs}} for the face corresponding to Sample 1 in Figure~\ref{fig:training_mani} for models CNN+lrn and ResNet respectively. Notice significant differences in the {\bf{xGEMs}} and their trajectories even at comparable overall performance.

\end{document}